# Management of Uncertainty in the Multi-Level Monitoring and Diagnosis of the Time of Flight Scintillation Array

**Robert K. Paasch**
Department of Mechanical Engineering
Oregon State University
Corvallis, OR 97331
paasch@kepler.me.orst.edu

**Alice M. Agogino**
Department of Mechanical Engineering
University of California
Berkeley, CA 94720
aagogino@euler.berkeley.edu

## Abstract

We present a general architecture for the monitoring and diagnosis of large scale sensor-based systems with real time diagnostic constraints. This architecture is multi-leveled, combining a single monitoring level based on statistical methods with two model-based diagnostic levels. At each level, sources of uncertainty are identified, and integrated methodologies for uncertainty management are developed. The general architecture was applied to the monitoring and diagnosis of a specific nuclear physics detector at Lawrence Berkeley National Laboratory that contained approximately 5000 components and produced over 500 channels of output data. The general architecture is scalable, and work is ongoing to apply it to detector systems one and two orders of magnitude more complex.

## 1 INTRODUCTION

The Time of Flight Scintillation Array is a sub-atomic particle detector used at Lawrence Berkeley National Laboratory (LBL) for studies in relativistic heavy-ion physics. With operating costs for these experiments exceeding $60,000 per hour, it is important that this detector operate correctly, and that any failures be identified and remediated as quickly as possible. But with approximately 5000 components, and over 500 output channels producing a Mbyte of data every 10 seconds, the monitoring and diagnosis of this detector system is difficult for human operators to accomplish within the real time constraints.

While the automation of the monitoring and diagnostic process was desirable, the scale of the Time of Flight Scintillation Array combined with real time constraints and uncertain system relationships presented a particular challenge to automated monitoring and diagnosis. The scale of the detector requires the average time spent monitoring an individual output channel (or probe) and diagnosing an individual component be small in order to meet the real time constraints. The uncertain nature of the relationship between the output data and the component states indicates an evidential reasoning approach, which may be contrary to the small average time requirement if applied to every component.

We present a general multi-level architecture developed to efficiently and non-deterministically diagnose large scale sensor-based systems in real time, and describe the implementation of this architecture to diagnose the Time of Flight Scintillation Array. The diagnostic system, the TOF Validation System, combines a single monitoring level with two model-based diagnostic reasoning levels to provide both efficiency and robustness in the face of uncertainty. The system consists of: 1) a statistical monitoring level using traditional chi-squared testing on data samples; 2) a model-based reasoning level operating on Boolean channel states (OK and BAD) with a data base of connectivity information to produce an ordered, reduced set of suspect components; and 3) a model-based reasoning level that extracts considerably more information from the data channels, and uses qualitative behavioral system information operating on the reduced set of suspect components to produce an evidential mapping to individual component failure state beliefs. The system systematically extracts more information from a reduced set of channels to provide a more detailed diagnosis. The system was developed to be scalable to diagnose similar detector systems two orders of magnitude more complex, and ongoing work is directed toward extending the system for use on a Time Projection Chamber at LBL and for possible use on the Superconducting Supercollider.

We believe the TOF Validation System is unique among sensor-based diagnostic expert systems developed to date, first due to the scale and complexity of the TOF Scintillation Array, second due to the multi-level nature of it's diagnostic reasoning, and third in the consideration and management of uncertainty at all levels of reasoning. The general methodology should be applicable to a broad range of monitoring and diagnostic problems where at least a qualitative model of the system is known.

In the following sections we will discuss other related research, describe the TOF Scintillation Array, present the general diagnostic system architecture, then discuss the management of uncertainty at the individual levels. We also describe the implementation of the generalized architecture and methodologies to develop the TOF Validation System.

## 2 BACKGROUND LITERATURE

In the development of an automated real-time sensor-based multi-level monitoring and diagnostic system, a broad variety of research and implementation issues must be addressed and so there exists a wealth of prior research work that is relevant to this research. This work builds on that of Agogino (1988a, b) and Rege (1986) in the area of real time sensor-based diagnostic expert systems. Rege describes IDES (Influence Diagram Expert System) and an application to a simple sensor-based pump diagnostic problem. Agogino describes the application of IDES to the real time sensor-based diagnosis of a milling machine and Ramamurthy (1990) to drilling applications.

Our view of multi-level diagnostic knowledge is similar to that of Milne (1985, 1987), who states that the diagnostic knowledge can exist at one or more of four levels: compiled, functional, behavioral and structural . Although compiled diagnostic knowledge systems often implicitly contain knowledge about structure and/or function, it is the explicit use of structural, behavioral or functional knowledge that delineates those diagnostic knowledge systems. Milne states that "the basic knowledge required for diagnosis is the set of malfunctions and relations between the observations and malfunctions". A compiled knowledge diagnostic system is a system that has this knowledge explicitly given to it. A structural diagnostic system is a system that is explicitly given structural or connectivity information, and likewise functional and behavioral diagnostic systems are explicitly given functional or behavioral information. While Milne uses multiple levels to categorize knowledge representation, we have explicitly used these same levels to describe a diagnostic architecture.

Chandrasekaran (1983), Davis (1983), Fink (1985a, b, 1987), and Scarl (1987) all address compiled verses deep (structural, behavioral and functional) knowledge based diagnosis. The research by Fink is particularly relevant to this research in that she describes a system that combines knowledge from more than one level, although the application described in that research is several orders of magnitude simpler than the TOF Scintillation Array. Scarl makes explicit use of both structure and function in the development of the diagnostic expert system for the space shuttle described in Scarl (1985).

When a diagnostic reasoning system depends on sensor information, the problem of sensor validation must be addressed. Chandrasekaran (1988) addresses sensor validation in compiled systems, and introduces a "meta-level" of compiled information that constitutes a level of redundancy based on expectations derived during diagnosis. Scarl (1987) shows that with the use of knowledge of structure and function one is able to regard sensor validation as a subset of the more general diagnostic process and therefore validate sensors the same as diagnosing any other component, an observation that we were able to confirm in this research.

## 3 TIME OF FLIGHT SCINTILLATION ARRAY

The TOF Scintillation Array is part of the Heavy Ion Superconducting Spectrometer (HISS) experiments at Lawrence Berkeley Laboratory. The TOF Scintillation Array consists of 136 plastic scintillation slats, 272 photomultiplier tubes, and associated electronics. As atomic particles produced by an experimental "event" pass through a particular slat, light photons are produced. The duration of an event is measured in nanoseconds, and the associated photons migrate to the ends of the slat and the produce electrical signals in the two photomultiplier tubes that are amplified 106 times. The signal is split and sent to both an Analog to Digital Converter (ADC) and a Time to Digital Converter (TDC). Digital outputs from the ADCs and TDCs are sent directly to magnetic tape, although sampling is possible. Because of the tremendous amount of data produced by and the scale of the TOF Scintillation Array, it can be difficult to monitor and diagnose even for an expert in the Wall's operation. Proposed detector systems for the Superconducting Supercollider would be humanly impossible to monitor and diagnose without assistance.

## 4 SYSTEM ARCHITECTURE

Difficulty in diagnosing large scale systems comes from a combination of both the complexity of the system and from the external constraints on the time allowed for diagnosis. Diagnostic systems using compiled knowledge (such as heuristic, rule-based systems) are difficult to implement for large scale systems due to the difficulties in acquiring and implementing large numbers of rules, rule conflict resolution, and difficulties in updating and appending the rule base. Also, as the number of rules increases, diagnostic time can increase. If the time to diagnose a failure is of little or no concern, then large, complex rule based systems may be a viable option, but this is not the case for this application.

When a system model is available, model based diagnostic systems can provide increased flexibility in handling system changes. But if the reasoning strategies applied to the entire system are too complex, model based diagnostic systems may also require too much time when diagnosing large scale systems.

The generalizable architecture we propose for large scale sensor-based systems closely mimics the methods by which an expert might diagnose a problem: first a quick look at all the probe outputs to get a general idea of what components might have failed, then a more detailed look at specific probes to determine the exact component and the specific failure type. This architecture consist of four main functional modules as shown in Figure 1.

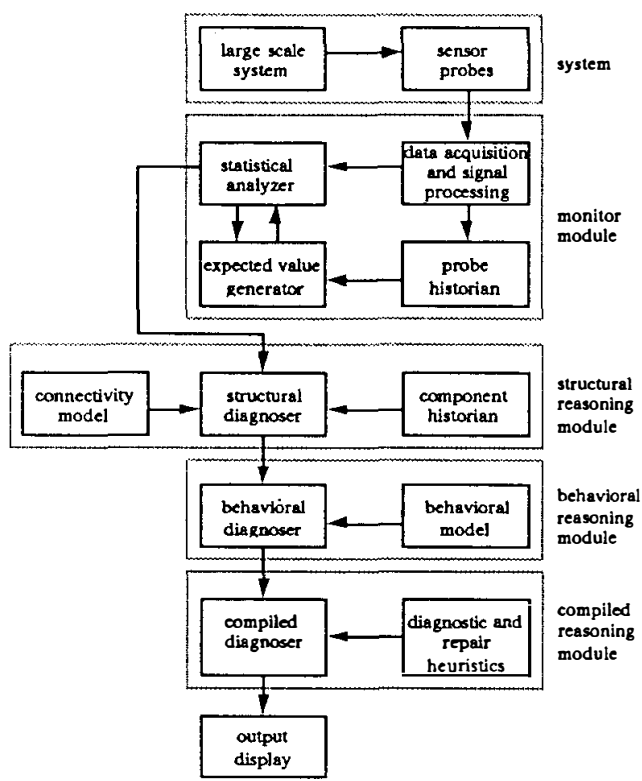

Figure 1: Large Scale System Diagnostic Architecture

A monitoring module would sample the probe (sensor) outputs, comparing these outputs to historical values. When the monitoring module detects an abnormality, the structural reasoning module is alerted. Using a connectivity model of the system and the probe output comparisons from the statistical analyzer, the structural reasoning module performs a "rough cut" diagnosis that produces a short list of suspect components. This short list is sent to the behavioral reasoning module. Using the short list of suspect components, this module retrieves the original probe outputs of all probes related to these components from the statistical analyzer, and using a functional model of the system produces a list of specific failure types for specific suspect components, ordered by degree of belief. A compiled reasoning model provides assistance in interpretation and repair.

Possible applications of this architecture include sensor validation in process control, fault location in complex electomechanical systems and control of food or chemical proccessing. We have applied this architecture to the monitoring and diagnosis of the Time of Flight Scintillation Array. In this implementation of the architecture, a data acquisition and signal processing unit compiles averages for ADC and TDC values for 1000 events, and a statistical analyzer compares these averages to archived values from the expected value generator with a chi-squared test. Probes that fail the test are flagged as "BAD", and the structural reasoning module started. This module assumes structural dependencies among the components based on connectivity (structure).

The term "structural dependency" in this context means that the effective state of a component is dependent on a preceding component. This dependency could occur when a component provides a data output that is the data input to a succeeding component, or if a component cannot operate if the preceding component has failed (such as when the preceding component supplies power to the succeeding component).

From the list of "BAD" probes from the statistical analyzer, and connectivity knowledge stored in the connectivity model, the structural diagnoser produces a short list of suspect components. The behavioral diagnosis module is then started with the list of suspect components. For each suspect component, this module solicits additional information from the statistical analyzer about trending of data from select groups of probes affected by the components under consideration, and compares those trends to expected trending (from the behavioral model) for individual failure types for the specific class of component using evidential algorithms. The final list of suspect components, with the types of failure, is presently sent directly to an output display. The compiled reasoning module has not been implemented.

## 5  MANAGEMENT OF UNCERTAINTY AT THE MONITORING LEVEL

The monitoring module addresses the question of whether the Time of Flight Scintillation Array is functioning correctly or not. Uncertainty exists at this level due to the highly stochastic nature of particle physics experiments.

In this case the monitoring process is abetted by the tremendous amount of data produced by the experiment. Samples of statistically significant size (typically 1000 events) can be collected rapidly. These samples are compared to expected norms (10 samples of 1000 events collected with the system assumed to be functioning correctly) using standard chi-square testing. These tests establish whether an individual data probe is OK or BAD. Statistical analysis validated the suitability of the chi-square test for determining the state of a probe (Paasch 1990).

# 6 MANAGEMENT OF UNCERTAINTY AT THE STRUCTURAL REASONING LEVEL

The structural reasoning module has the responsibility for reducing the set of suspect components from a set of all components in the system to a set that can be efficiently diagnosed by the behavioral reasoning module within the real time constraints. The largest reduction in the suspect set occurs in this module. In the TOF Validation System, this module reduces the suspect component set from 5000 components to approximately 10. Detailed information on this module is included in Hall (1989).

The structural reasoning module assumes two possible component states (OK and BAD), and two possible observations per observation probe (OK and BAD). We justify this simplification of component states on the basis of highest common denominator. All components can use OK and BAD, while failure type may be component specific. Also, in the structural reasoning module a rough diagnosis is acceptable, and the lumping of components into either OK or BAD is sufficient. As for the observations per probe, OK or BAD may be all the information available from some extracted features such as the chi-square test used by the monitoring module.

Uncertainty in the probe state is addressed at the monitoring level by statistical methods. Uncertainty in the component state is addressed to some degree at this level, and again at the behavioral reasoning level. At this level we deterministically assume that if a component has all dependent probes in state OK then that component is in state OK and is removed as a suspect. This assumption is justified by the statistical significance of the sample size used in the monitoring module.

If a component has one or more dependent probes in state BAD, then that component can be included as a suspect. Uncertainty about the individual states of the components on the reduced suspect list can be handled by ranking the suspect components by the BAD probe to total probe ratio: a component with four probes dependent upon it and three of those probes assumed BAD would rank ahead of a component with eight dependent probes of which four are BAD.. Multiple component failures would increase the size of the suspect list. The possibility of multiple failures would decrease with the time to collect a sample.

If two or more suspect components have identical signatures (affect the same probes) we cannot discriminate between component failures, an ambiguity would exist, and thus would present a failure class. For the TOF Scintillation Array the worst case is three ambiguous states (i.e. single components failures). Compiled information in the form of prior probabilities can handle this ambiguity, or a secondary diagnosis (detailed in the next section) can be performed.

# 7 MANAGEMENT OF UNCERTAINTY AT THE BEHAVIORAL REASONING LEVEL

Although the structural reasoning module efficiently produces a limited set of suspect components and can order that list, a more detailed diagnosis may be desired. The structural reasoning module is sensitive to the breakpoint value between BAD and OK channels, in the best of circumstances it may produce a suspect component list with some ambiguity, and it provides no information the specific type of component failure. The behavioral reasoning module was incorporated into the general architecture to address these problems. This module by itself would be adequate for the diagnosis of small scale systems, but is too computationally inefficient to operate on a large set of suspect components in real time.

With the apparent complexity of the system greatly reduced, the behavior module can operate in numerous ways. The behavioral reasoning architecture we present incorporates a behavioral model to produce expected observation values for the different failure types, a methodology to compare expected values to actual values, and a methodology to relate that comparison to the system states. In the TOF Validation System, the later two methodologies are implemented in the behavioral diagnoser.

The behavioral model can exist on many different levels. The compilation of historical data, keeping track of failures by type, would result in a compiled behavioral model relating system state directly to observables. At this level the knowledge would be considered shallow: compiled knowledge of system operation as a whole, with little or no knowledge of deeper system operation. At the other extreme would be the deep analytical model: every component in the system, and every relationship between components is modeled analytically, with a resultant complex math-ematical equation for the system rigorously relating system state to observables. Between these two extremes there are a number of possibilities, including qualitative models, data models based on experiential, first principle, compiled knowledge, and hybrid models.

Numerical comparison methodologies such as chi-square or Z distribution can work for numerical observations, and Boolean sensor outputs can be compared directly with expected outputs on a match/no-match basis, with the result mapped directly to increase or decrease a belief.

Relational methodologies would generally involve some sort of mapping from comparison value to evidential value. This mapping could be as simple as a table look up, or could involve qualitative or quantitative relational algorithms.

One possible feature of a behavioral reasoning module, implemented in the TOF Validation System, is the continuous mapping made between data values and hypothesis belief. The structural reasoning module assumes that belief is discrete, as this is a Boolean mapping. For example, if probability were used, $p(t = BAD|y < limit) = 0$ and $p(t = BAD|y > limit) = 1$, where t is a probe state, y is the probe data value, and "limit" is the established breakpoint between OK and BAD. But the argument can be made that the uncertainty mapping in this and other cases is continuous. Intuitively, $p(t = BAD|y)$ might increase as y increases. More importantly, we can relate the individual component failure states, $s_i$, to the probe values. For each evidence type for each hypothesis, an algorithm could be found that relates $p(s_i|y)$ to the value of y. A plot of such an algorithm is shown in Figure 2. In this case the likelihood of the hypothesis, H, increases as the value of the feature, x, increases, but the likelihood might also decrease with an increasing feature value, or it might increase with any feature change, or be related in other ways. The tradeoff to using a continuous evidential approach may be computational complexity and the inability to invert.

Evidential methodologies are used to combine the evidence produced by the relational algorithms, and could be based on Bayesian probabilities, Certainty Factors and Dempster-Shafer, at the preference of the expert or system developers. Each individual observation contributes an individual belief or likelihood for each system state hypothesis, these are then combined to give a final belief or likelihood for each system state hypothesis. The final belief or likelihood for each system state could then be used on either an absolute or relative basis.

The behavioral reasoning module as implemented in the TOF Validation System uses numerical, relational and evidential sub-levels. A qualitative behavioral model is based on the experts first principle and expe-riential knowledge of the TOF Scintillation Array. To compare current with expected data features a simple Z distribution comparison ( z= (x - xavg)/sx ) is used. The behavioral diagnoser then uses four continuous quantitative relational algorithms, one for each expected direction of data trending (increasing, decreasing, either direction, and no change) to relate the data to evidential belief. These algorithms are modified for individual component failure state/ data feature relationships by parameters to reflect the desired evidential weighting, slope and cutoff value as determined by the expert. The algorithms can be made to behave in a nearly discrete manner when appropriate. At the preference of the expert, the present implementation uses certainty factors as an evidential methodology. The conversion to Bayesian probabilities is straightforward with the addition of the assessment of prior probabilities (Heckerman 1986).

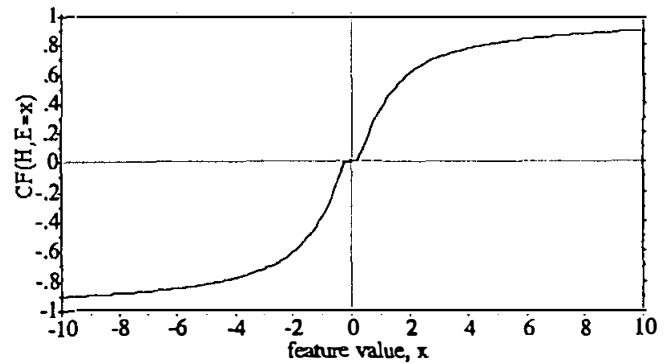

Figure 2: Increase in Belief Associated With Evidence Verses Strength of the Evidence

## 8 IMPLEMENTATION

The architecture was implemented by the development and testing of individual modules that were then combined to form the TOF validation System. The data acquisition and signal processing unit was coded in a combination of FORTRAN and C. All other modules were coded in C. The system runs on the HISS experiment's VAX computers operating under VMS. A more detailed description of the implementation and use of the validation system may be found in Hall (1989), Olson (1990), and Paasch (1988, 1990).

## 9 SUMMARY

This paper presents a multi-level diagnostic architecture particularly well suited for the diagnosis of large scale systems with real time constraints. While the idea of multi-level diagnosis is not new, we believe this system is unique in that the inherent complexity and scale of the application required that multiple levels be used in this case to achieve the requisite efficiency while maintaining diagnostic detail and the ability to manage uncertainty.

In implementing a multi-level diagnostic system, we have found that the individual diagnostic levels have strengths and weaknesses that are complementary and generalizable. Structural reasoning systems can be fairly robust, and can be designed in a generic manner so that changes in the structure of the components are reflected as changes in a mapping database. Due to the inherent ambiguity, structural reasoning appears to best lend itself to a rough, first pass, preliminary diagnosis. Uncertainty at this level exists as ambiguity between system states and as uncertainty in the probe feature value to symbolic value mapping.

Behavioral reasoning systems, especially analytic systems derived from first and second engineering principles and incorporating structural knowledge, can be very robust. The construction of a reasoning system from behavioral knowledge can be difficult, however, due to incomplete information on the operation of a complex system and the difficulty in rigorously mapping from evidence to belief. Uncertainty exists in this mapping, and in the accuracy of the system model. Behavioral reasoning systems might need to incorporated various levels of compiled behavioral information, making the design of a general framework more difficult.

We do not feel that compiled systems by themselves are appropriate for the diagnosis of real time large scale systems because of complexity and inflexibility. However, compiled reasoning levels have been considered for this application to interpret the output from the behavioral reasoning module and to guide the repair process.

As the application of diagnostic reasoning systems to more complex systems becomes possible, it becomes apparent that these reasoning systems will incorporate a preprocessing level combined with multiple levels of diagnostic reasoning: structural, behavioral, and compiled. Each level has specific strengths and weaknesses, and in this research a combination has proven to provide the most intelligent, efficient, flexible, and robust system for diagnosing large scale real-time systems like the TOF Scintillation Array and proposed detectors for the Superconducting Supercollider.

## Acknowledgements

The authors would like to thank Dennis Hall, Bill Greiman and Doug Olson from Lawrence Berkeley National Laboratory for their guidance and consultation throughout this research. The authors greatly appreciate the support of the Director, Office of Energy Research, Scientific Computing Staff, of the United States Department of Energy, under Contract Number DE-AC03-76SF00098, and by the National Science Foundation, under grant DMC-8451622.